\pdfoutput=1
\documentclass[letterpaper]{article}
\usepackage{aaai2027}
\usepackage[hyphens]{url}
\usepackage{graphicx}
\usepackage{booktabs}
\usepackage{multirow}
\usepackage{natbib}
\usepackage{caption}
\usepackage{hyphenat}
\usepackage{enumitem}
\usepackage{amsmath}
\setlist[itemize]{topsep=2pt, partopsep=0pt, itemsep=1pt}
\setlist[enumerate]{topsep=2pt, partopsep=0pt, itemsep=1pt}
\usepackage{amsmath}
\usepackage{amssymb}
\usepackage{bm}
\usepackage{soul}
\definecolor{bestbg}{RGB}{255,220,220}

\usepackage{xcolor}
\usepackage{colortbl}
 
\nocopyright
\title{QIRF: Quantum-Inspired Non-Orthogonal Function-Space Compression for 3D Gaussian Splatting}
\author{
\textbf{Shizeng Jiang}\thanks{First Author}\textsuperscript{1,2},
\textbf{Hao Zhang}\textsuperscript{1,2}\thanks{Corresponding Author},
\textbf{Xuerui Ma}\textsuperscript{3,4},
\textbf{Ying Hu}\textsuperscript{3,4},
\textbf{Tao Zhang}\textsuperscript{1,2,5}\thanks{Corresponding Author}
}
\affiliations{
\textsuperscript{1,2,5} School of Electronic Information, Central South University, China \\
\textsuperscript{3,4} Hunan Malanshan Audio and Video Laboratory, China
}

\begin{document}
\maketitle

\begin{abstract}
3D Gaussian Splatting (3DGS) achieves high-quality real-time rendering by representing a scene with a large collection of anisotropic Gaussian primitives. However, complex scenes often require millions of Gaussians, resulting in substantial storage and rendering costs. Existing compression methods mainly reduce redundancy through primitive-wise pruning, attribute quantization, clustering, or neural coding, while redundancy caused by strongly overlapping and non-orthogonal Gaussian basis functions remains largely unexplored. We present QIRF, a quantum-inspired non-orthogonal function-space compression for 3D gaussian splatting. QIRF models neighboring Gaussian primitives as a local non-orthogonal basis and formulates primitive reduction as a subspace-aware selection problem. Specifically, an analytic Gaussian overlap matrix and a radiance-response density matrix are constructed to characterize functional redundancy and rendering relevance. Generalized eigendecomposition is then used to identify the dominant local subspace and select representative Gaussian primitives. An RRDM-based response model and detail-aware safeguarding further preserve visually important high-frequency structures under aggressive pruning. Experiments on 13 scenes from Mip-NeRF 360, Tanks\&Temples, and Deep Blending show that QIRF reduces the Gaussian count and raw PLY storage by $71.7\%$ on average, corresponding to approximately $3.54\times$ compression, while maintaining reconstruction quality comparable to 3DGS and achieving a marginal average PSNR improvement of $0.10$ dB. QIRF also improves the average rendering speed over 3DGS by $34.3\%$. These results suggest that non-orthogonal function-space redundancy is an important yet underexplored source of representational redundancy in explicit Gaussian radiance fields.
\end{abstract}

\section{Introduction}
Novel-view synthesis (NVS) aims to reconstruct a scene from calibrated images and render photorealistic views from unseen camera poses. Neural Radiance Fields (NeRFs) \citep{mildenhall2021nerf} achieve remarkable rendering quality but typically require expensive volumetric sampling. In contrast, 3D Gaussian Splatting (3DGS) \citep{kerbl20233dgs} represents a scene using explicit anisotropic Gaussian primitives and renders them through differentiable rasterization, enabling efficient optimization and high-quality real-time rendering. However, efficient rasterization does not necessarily imply a compact scene representation. Adaptive densification repeatedly introduces new primitives into under-reconstructed regions \citep{lee2024compact3dgs,fan2023lightgaussian}, causing complex scenes to contain millions of Gaussians and occupy hundreds of megabytes or even more than one gigabyte in raw PLY format. This redundancy increases storage requirements, GPU memory consumption, transmission bandwidth, and rasterization cost.

Existing compact 3DGS methods address this problem through primitive pruning \citep{fan2023lightgaussian,wang2024endtoendrd}, attribute quantization \citep{fan2023lightgaussian,chen2024hac}, codebook learning \citep{lee2024compact3dgs}, anchor-based prediction \citep{lu2024scaffoldgs,chen2024hac}, and entropy coding \citep{chen2024hac,wang2024endtoendrd}. In particular, pruning-based approaches generally assign an independent importance score to each Gaussian according to opacity, visibility, gradient magnitude, or rendering contribution. Although effective, such primitive-wise criteria do not explicitly model the functional redundancy caused by strongly overlapping anisotropic Gaussians. Two Gaussians may both receive high individual importance scores while spanning nearly the same local scene function. Therefore, determining whether a primitive is individually important is not equivalent to determining whether it contributes a unique direction to the underlying radiance-field representation.

Motivated by this observation, we view a 3DGS scene as a function space spanned by non-orthogonal Gaussian basis functions and formulate compact scene representation as a local basis-reduction problem. Inspired by non-orthogonal basis analysis in quantum chemistry \citep{hall1951molecular,roothaan1951new,lowdin1955quantum}, we introduce QIRF, a quantum-inspired non-orthogonal function-space compression for 3D gaussian splatting. Within each spatial block, QIRF constructs an analytic overlap matrix to characterize correlations among neighboring Gaussian basis functions and a radiance-response density matrix to describe their accumulated rendering relevance. Generalized eigendecomposition is then used to identify the dominant local response subspace, from which representative Gaussian primitives are selected according to their participation in the retained modes. An RRDM-based response model further integrates multiple rendering signals, while detail-aware safeguarding preserves visually important primitives around thin structures, high-frequency textures, and occlusion boundaries. Experiments on 13 scenes from Mip-NeRF 360, Tanks\&Temples, and Deep Blending show that QIRF removes $71.7\%$ of Gaussian primitives on average and achieves approximately $3.54\times$ raw PLY compression while maintaining reconstruction quality comparable to 3DGS, with a marginal average PSNR improvement of $0.10$ dB. QIRF also improves the average rendering speed over 3DGS by $34.3\%$.

Our contributions are summarized as follows:
\begin{enumerate}
    \item We formulate compact 3D Gaussian Splatting as a local non-orthogonal basis-reduction problem and explicitly model functional redundancy among overlapping anisotropic Gaussian primitives.

    \item We introduce a subspace-aware primitive selection method that combines an analytic Gaussian overlap matrix with a radiance-response density matrix through generalized eigendecomposition.

    \item We develop an RRDM-based response model and a detail-aware safeguarding strategy to preserve visually important structures under aggressive primitive reduction.

    \item Experiments on 13 benchmark scenes demonstrate that QIRF removes $71.7\%$ of Gaussian primitives on average, achieves approximately $3.54\times$ raw PLY compression, and improves the average rendering speed over 3DGS by $34.3\%$, while maintaining comparable reconstruction quality.
\end{enumerate}

\section{Related Work}

\subsection{Neural Radiance Fields and 3D Gaussian Splatting}
Neural Radiance Fields (NeRFs) represent scenes as continuous neural functions and synthesize novel views through differentiable volume rendering \citep{mildenhall2021nerf}. Subsequent methods, such as mip-NeRF, mip-NeRF 360, and Zip-NeRF, improve anti-aliasing, rendering quality, and robustness in complex or unbounded scenes \citep{barron2021mipnerf,barron2022mipnerf360,barron2023zipnerf}. However, NeRF-based approaches generally rely on dense sampling along camera rays and repeated network evaluations, which limits training and rendering efficiency. To reduce this computational cost, explicit and structured radiance-field representations have been widely explored. PlenOctrees and SNeRG convert learned radiance fields into structures suitable for fast rendering \citep{yu2021plenoctrees,hedman2021baking}, while Instant-NGP accelerates optimization through multiresolution hash-grid encoding \citep{muller2022instantngp}. Other methods improve efficiency using sparse voxel grids, tensor decomposition, or planar factorization \citep{fridovich2022plenoxels,sun2022directvoxel,chen2022tensorf,fridovich2023kplanes}.

3D Gaussian Splatting (3DGS) \citep{kerbl20233dgs} represents a scene using explicit anisotropic Gaussian primitives and renders them through differentiable rasterization. This formulation enables efficient optimization and high-quality real-time rendering without dense volumetric sampling. However, the adaptive densification process of 3DGS may generate millions of Gaussian primitives in complex scenes. The resulting over-parameterized representation increases storage requirements, GPU memory consumption, transmission bandwidth, and rasterization cost, motivating extensive research on compact and efficient 3D Gaussian representations.
\subsection{Compact 3D Gaussian Representations}
The explicit representation of 3D Gaussian Splatting (3DGS) has motivated extensive research on compact scene representations. Existing methods mainly reduce redundancy through \emph{primitive reduction} and \emph{attribute compression}.

Primitive-reduction methods remove visually insignificant or redundant Gaussians according to opacity, visibility, gradients, or rendering contribution. LightGaussian performs importance-based pruning followed by lightweight recovery optimization to improve reconstruction quality after pruning \citep{fan2023lightgaussian}. More recent approaches, including MaskGaussian \citep{liu2025maskgaussian} and RadSplat \citep{niemeyer2025radsplat}, further improve primitive selection using structured masking or rendering-aware importance estimation, enabling more compact Gaussian representations while preserving rendering quality. Although effective, these methods generally evaluate the importance of each Gaussian independently using scalar scores or learned masks, without explicitly modeling the functional redundancy among overlapping Gaussian basis functions.

Attribute-compression methods instead focus on reducing the storage cost of retained primitives. Compact3DGS compresses Gaussian attributes through vector quantization, learned codebooks, and entropy coding while preserving reconstruction quality \citep{lee2024compact3dgs}. Such methods significantly reduce model size by optimizing parameter representations and coding schemes, yet they still treat Gaussian primitives as independent entities and do not explicitly exploit the function-space correlation induced by overlapping Gaussian bases.

Unlike previous approaches, QIRF performs compression from a function-space perspective. Rather than assigning independent importance scores or compressing only Gaussian attributes, it constructs a local non-orthogonal basis using neighboring Gaussians and identifies dominant scene subspaces through generalized natural scene orbital analysis. This enables QIRF to remove functionally redundant Gaussian basis functions before representation optimization, making it complementary to existing pruning- and attribute-compression-based methods.

{Generalized Eigenanalysis in Non-Orthogonal Function Spaces}
Representations constructed from overlapping basis functions are commonly analyzed through an overlap matrix that defines the metric of a non-orthogonal basis space. In quantum chemistry, the Roothaan--Hall formulation solves a generalized eigenvalue problem to account for correlations among non-orthogonal atomic orbitals \citep{roothaan1951new,hall1951molecular}. Natural-orbital analysis further diagonalizes a density operator and uses the resulting occupation values to identify compact low-rank subspaces \citep{lowdin1955quantum,lowdin1956natural}. Similar generalized eigendecomposition techniques have also been explored in numerical linear algebra and learning problems involving non-orthogonal representations \citep{golub2013matrix,saad2011numerical,huggins2020nonorthogonal}.These studies suggest that generalized eigendecomposition provides a principled way to identify dominant subspaces when basis functions are correlated and non-orthogonal. Such a perspective has received little attention in compact 3D Gaussian Splatting, although neighboring anisotropic Gaussian primitives also exhibit substantial spatial overlap and functional correlation. Consequently, representing a scene as a collection of independent Gaussian primitives may overlook redundancy existing in the underlying local function space.

Unlike standard principal component analysis (PCA), which assumes an Euclidean inner product and orthogonal basis vectors \citep{jolliffe2016principal}, generalized eigendecomposition explicitly incorporates the overlap matrix as the metric of the local basis space \citep{roothaan1951new,hall1951molecular}. Consequently, the extracted dominant modes jointly account for both rendering responses and basis redundancy, making the formulation more suitable for compact representations built from overlapping Gaussian primitives.

Building upon this observation, QIRF models neighboring Gaussian primitives as a local non-orthogonal basis and formulates Gaussian selection as a generalized eigenvalue problem. The resulting dominant subspace provides a function-space criterion for identifying representative Gaussian primitives, which differs fundamentally from independent importance scoring and attribute-level compression.

\section{Method}
\subsection{Background}

3D Gaussian Splatting (3DGS) represents a scene using explicit anisotropic Gaussian primitives. Each Gaussian is parameterized by its center position $\boldsymbol{\mu}_i$, covariance matrix $\boldsymbol{\Sigma}_i$, opacity $\alpha_i$, and view-dependent appearance coefficients $\mathbf{c}_i$. Its continuous density is

\begin{equation}
G_i(\mathbf{x})=
\exp\!\left(
-\frac12
(\mathbf{x}-\boldsymbol{\mu}_i)^{\!\top}
\boldsymbol{\Sigma}_i^{-1}
(\mathbf{x}-\boldsymbol{\mu}_i)
\right),
\label{eq:gauss-density}
\end{equation}

where $\mathbf{x}\in\mathbb{R}^3$. The covariance is parameterized as

\begin{equation}
\boldsymbol{\Sigma}_i
=
\mathbf{R}_i
\mathbf{S}_i
\mathbf{S}_i^{\top}
\mathbf{R}_i^{\top},
\label{eq:cov-decompose}
\end{equation}

where $\mathbf{R}_i$ and $\mathbf{S}_i$ denote the rotation and scaling matrices.

During rendering, Gaussians are projected onto the image plane and composited according to depth order,

\begin{equation}
\mathbf{C}(\mathbf{p})
=
\sum_{i\in\mathcal N(\mathbf p)}
T_i(\mathbf p)
\alpha_i(\mathbf p)
\mathbf c_i(\mathbf d),
\label{eq:color-comp}
\end{equation}

where

\begin{equation}
T_i(\mathbf p)
=
\prod_{j<i}
\left(1-\alpha_j(\mathbf p)\right).
\label{eq:transmittance}
\end{equation}

Although this explicit rasterization enables real-time rendering, adaptive densification often introduces millions of Gaussian primitives, resulting in substantial redundancy, storage overhead, and increased rasterization cost. Our objective is therefore to identify a compact subset of representative Gaussians while preserving rendering quality.

\subsection{Method Overview}
We present an overview of our proposed pipeline in Figure 1. Given an input 3DGS scene, we first partition Gaussian primitives into adaptive voxel blocks and model neighboring anisotropic Gaussians within each block as overlapping non-orthogonal basis functions. For each local block, QIRF constructs a Gaussian overlap matrix to measure basis redundancy and a Radiance-Response Density Matrix (RRDM) to describe rendering relevance. These matrices are then used for Generalized Natural Scene Orbital (GNSO) analysis, where a generalized eigenvalue problem identifies the dominant local function subspace and representative Gaussians are selected according to their participation in the retained modes. To preserve thin structures and high-frequency details, detail-aware safeguarding retains a small number of important inactive Gaussians before hard pruning. The selected primitives are finally fine-tuned without further densification, producing a compact 3DGS representation with reduced storage and maintained rendering quality.

\begin{figure*}[t]
\centering
\includegraphics[width=0.85\textwidth]{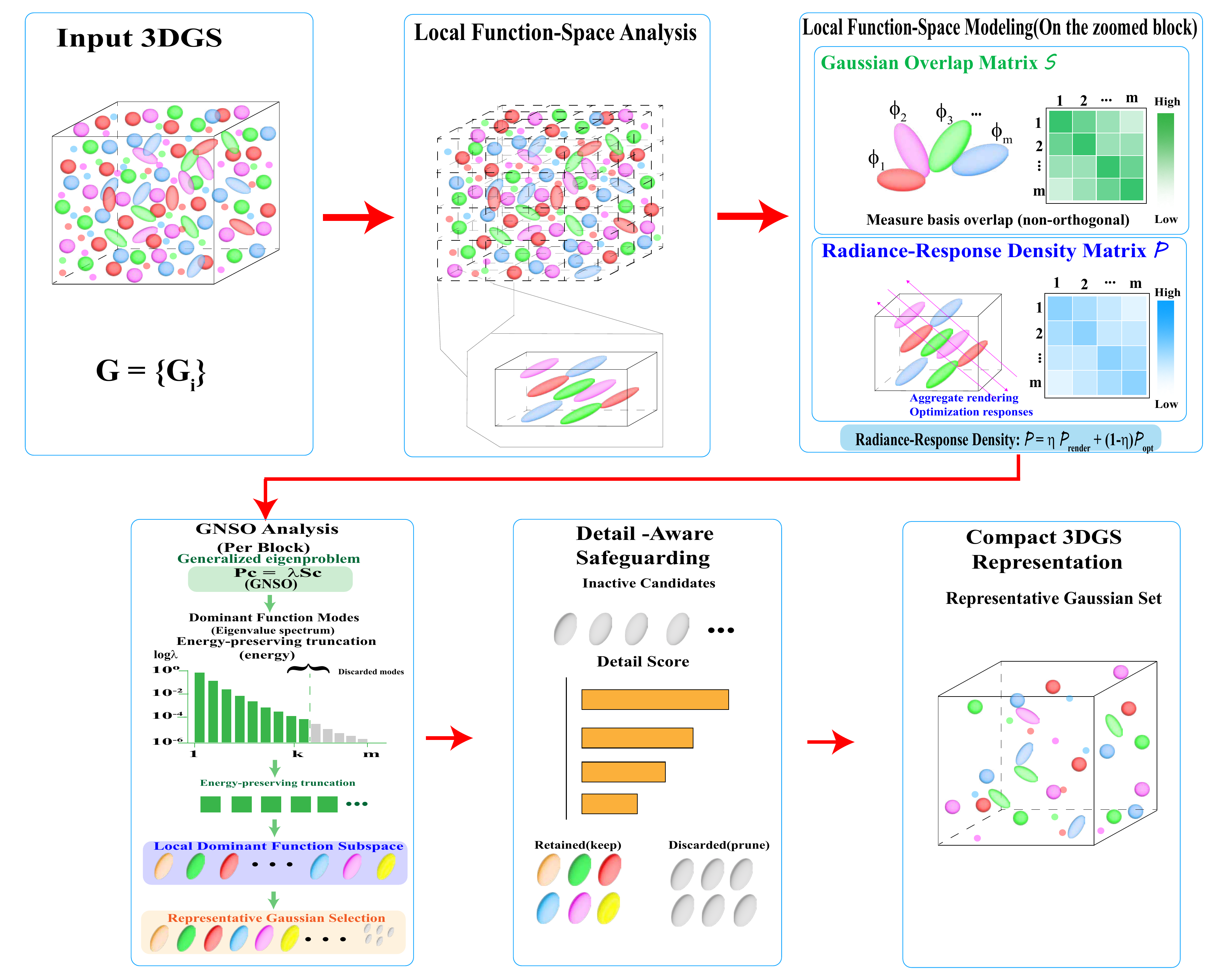}
\caption{Overview of QIRF. Local overlap and radiance-response density matrices guide GNSO-based subspace selection, while detail-aware safeguarding preserves important primitives for compact 3DGS representation.}
\label{fig:qirf_overview}
\end{figure*}

\begin{figure*}[t]
\centering
\includegraphics[height=8cm,width=0.8\textwidth]{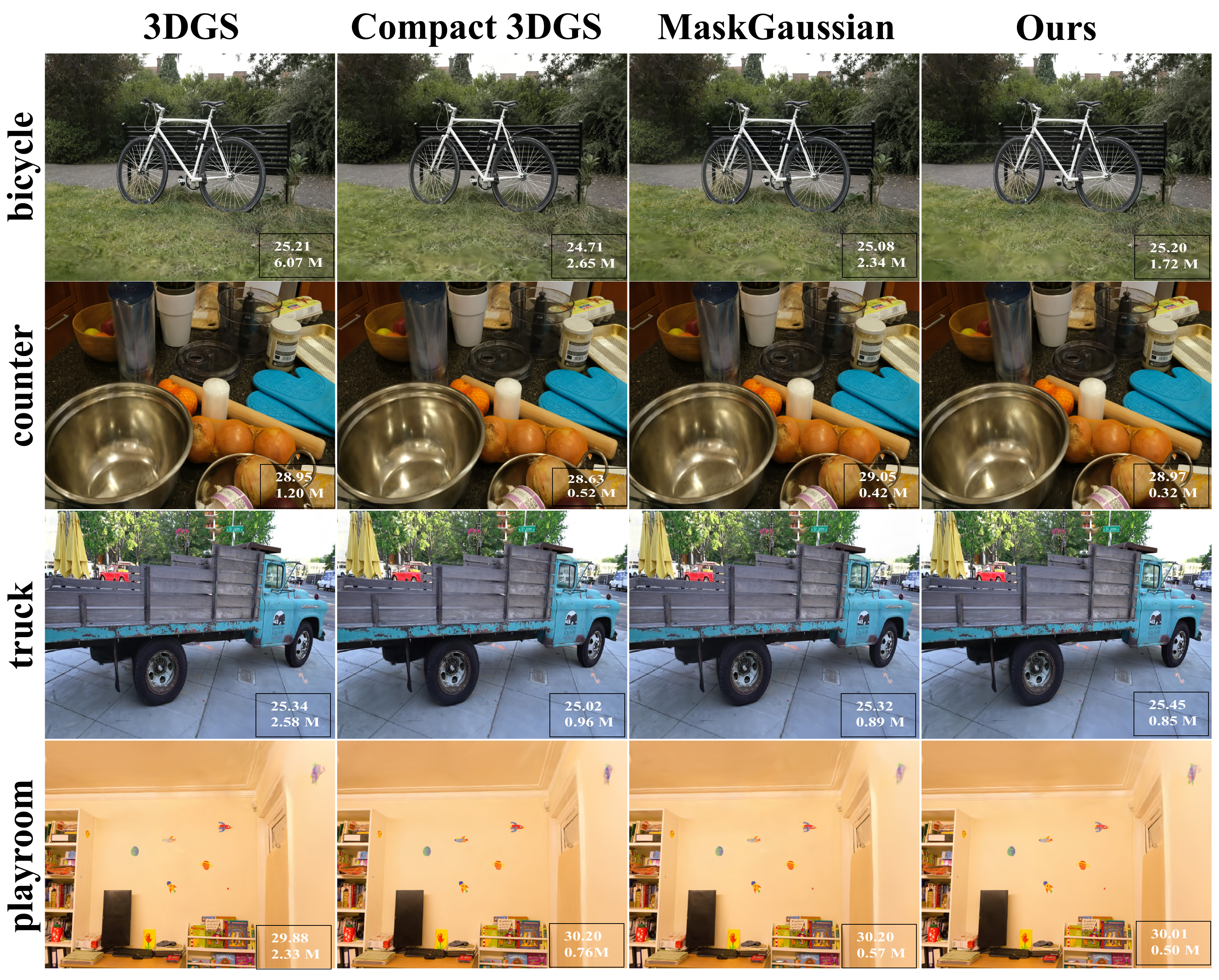}
\caption{Qualitative visual comparisons and quantitative results for QIRF, 3DGS, MaskGaussian, and Compact3DGS. We report PSNR and the number of Gaussians in millions.}
\label{fig:qual_compare}
\end{figure*}

\subsection{Problem Formulation and Adaptive Partitioning}
A 3DGS scene is represented by $N$ anisotropic Gaussian primitives
$\mathcal{G}=\{\mathcal{G}_i\}_{i=1}^{N}$, where each primitive is parameterized as
\begin{equation}
\mathcal{G}_i=
\left(
\boldsymbol{\mu}_i,
\boldsymbol{\Sigma}_i,
\alpha_i,
\mathbf{c}_i
\right).
\label{eq:gauss-param}
\end{equation}
Our goal is to identify a compact representative subset
\begin{equation}
\mathcal{A}\subseteq\{1,\ldots,N\},
\qquad |\mathcal{A}|\ll N,
\label{eq:active-set-def}
\end{equation}
while preserving novel-view rendering quality. Unlike primitive-wise pruning methods that evaluate each Gaussian independently, QIRF formulates compression as a local non-orthogonal basis-reduction problem. The objective is to retain primitives that span the dominant radiance-response subspace rather than simply removing Gaussians with low scalar importance scores.

Constructing global $N\times N$ correlation matrices is computationally infeasible for scenes containing millions of Gaussians. We therefore partition the scene into local voxel blocks and perform function-space analysis independently within each block. In practice, we use a scene-adaptive voxel size selected from a small candidate set according to the sparsity of the initial partition. Specifically, we first partition the scene with a base voxel size $v_0$ and compute the block sparsity ratio
\begin{equation}
\rho_{\mathrm{block}}
=
\frac{N_{\mathrm{blocks}}}{N_{\mathrm{Gaussians}}},
\label{eq:block-sparsity-ratio}
\end{equation}
where $N_{\mathrm{blocks}}$ denotes the number of non-empty blocks. A larger $\rho_{\mathrm{block}}$ indicates many low-occupancy blocks, for which a coarser voxel size is used to improve the stability of local subspace estimation. This adaptive choice depends only on training-scene geometry and does not use test-view metrics.

\subsection{Local Function-Space Modeling}

Within each voxel block, neighboring Gaussian primitives are interpreted as non-orthogonal spatial basis functions. For the $i$-th primitive, we denote the corresponding Gaussian basis by $\phi_i(\mathbf{x})$, following the anisotropic density defined in Eq.~\eqref{eq:gauss-density}. We first construct the local overlap matrix
\begin{equation}
S_{ij}
=
\int_{\mathbb{R}^3}
\phi_i(\mathbf{x})
\phi_j(\mathbf{x})
\,d\mathbf{x},
\label{eq:overlap-matrix-entry}
\end{equation}
which measures the functional similarity between two Gaussian bases. A large value of $S_{ij}$ indicates strong spatial overlap and potential functional redundancy. To improve numerical stability, we use the regularized metric
\begin{equation}
\widetilde{\mathbf{S}}
=
\mathbf{S}
+
\epsilon\mathbf{I},
\qquad
\epsilon=10^{-4}.
\label{eq:reg-overlap-matrix}
\end{equation}

Spatial overlap alone does not determine rendering relevance.
After densification stops, we accumulate for each Gaussian an
eight-dimensional response descriptor
$\bar{\mathbf r}_i\in\mathbb{R}^{8}$ using an exponential moving
average. The descriptor contains four rendering-response channels
(visibility, composited contribution, residual-weighted contribution,
and projected alpha mass) and four optimization-response channels
(gradient magnitudes of position, scale, opacity, and appearance).
The exact channel definitions and normalization are provided in the
supplementary material.

For a local block with $m$ Gaussians, we stack the normalized
rendering and optimization components into
$H_{\mathrm{r}},H_{\mathrm{o}}\in\mathbb{R}^{m\times4}$, respectively.
We then construct the Radiance-Response Density Matrix (RRDM) as
\begin{equation}
P=
\eta\frac{H_{\mathrm{r}}H_{\mathrm{r}}^\top}
{\operatorname{tr}(H_{\mathrm{r}}H_{\mathrm{r}}^\top)+\epsilon}
+
(1-\eta)\frac{H_{\mathrm{o}}H_{\mathrm{o}}^\top}
{\operatorname{tr}(H_{\mathrm{o}}H_{\mathrm{o}}^\top)+\epsilon},
\qquad \eta=0.75 .
\label{eq:rrdm}
\end{equation}
By construction, $P\in\mathbb{R}^{m\times m}$ is positive
semidefinite and encodes the similarity of accumulated rendering and
optimization responses between local Gaussians. Only the
$8$-dimensional per-Gaussian descriptors are stored during training;
the local RRDM is constructed once at the pruning iteration.

\subsection{Generalized Natural Scene Orbital Compression}
Given the overlap metric $\widetilde{\mathbf{S}}$ and the Radiance-Response Density Matrix $\mathbf{P}$, QIRF identifies the dominant local function modes by solving the generalized eigenvalue problem
\begin{equation}
\mathbf{P}\mathbf{u}_k
=
\lambda_k
\widetilde{\mathbf{S}}
\mathbf{u}_k,
\label{eq:general-eig-problem}
\end{equation}
where the eigenvectors $\mathbf{u}_k$ define the \textbf{Generalized Natural Scene Orbitals (GNSO)}, and the corresponding eigenvalues $\lambda_k$ measure the occupation strength of each local response mode.

Unlike conventional principal component analysis (PCA) or covariance decomposition, which assumes an Euclidean inner product, Eq.~\eqref{eq:general-eig-problem} explicitly incorporates the overlap matrix as the metric of the local basis space. Consequently, the extracted modes jointly account for rendering relevance and functional redundancy among neighboring Gaussian basis functions, making the formulation particularly suitable for compact explicit Gaussian representations.

After sorting the occupation values in descending order, we retain the smallest number of modes $r$ satisfying
\begin{equation}
\frac{
\sum_{k=1}^{r}\lambda_k
}{
\sum_{k=1}^{m}\lambda_k
}
\geq
\tau,
\qquad
\tau=0.95,
\label{eq:energy-preserve-thresh}
\end{equation}
which preserves the dominant local function subspace while discarding weak response modes.

The retained modes are mapped back to the original Gaussian primitives using an occupation-weighted participation score,
\begin{equation}
q_i
=
\sum_{k=1}^{r}
\lambda_k
|u_{ik}|^2,
\label{eq:gauss-participation-score}
\end{equation}
where a larger $q_i$ indicates a stronger contribution of the $i$-th Gaussian to the dominant local function subspace. To obtain a discrete primitive set, each block retains the top-$k_b$
participation scores, where
$k_b=\min\{m_b,\max(r_b,\lceil\kappa m_b\rceil)\}$ with
$\kappa=0.22$. Thus, $\tau$ determines the retained GNSO subspace,
whereas $\kappa$ controls the primitive budget. The complete
block-wise selection procedure is provided in the supplementary
material.

The resulting representative set forms the base compact representation, which is further refined by the following detail-aware safeguarding strategy.

\begin{table*}[!t]
\centering
\setlength{\abovecaptionskip}{4pt}
\setlength{\belowcaptionskip}{4pt}
\caption{Quantitative comparison of our method with prior work on three benchmark datasets. Results marked with $^*$ are cited from original papers; all other values are obtained via our reproduced experiments. \#GS denotes the number of Gaussians in millions. FPS denotes rendering speed. \colorbox{bestbg}{Best} results are highlighted in bold with light red background.}
\label{tab:tab1_basic_metric}
\setlength{\tabcolsep}{2.7pt}
\renewcommand{\arraystretch}{1.05}
\resizebox{\textwidth}{!}{
\begin{tabular}{lccccc ccccc ccccc}
\toprule
\multirow{2}{*}{Method}
& \multicolumn{5}{c}{Mip-NeRF 360}
& \multicolumn{5}{c}{Tanks\&Temples}
& \multicolumn{5}{c}{Deep Blending} \\
\cmidrule(lr){2-6} \cmidrule(lr){7-11} \cmidrule(lr){12-16}
& PSNR $\uparrow$ & SSIM $\uparrow$ & LPIPS $\downarrow$ & \#GS (M) $\downarrow$ & FPS $\uparrow$
& PSNR $\uparrow$ & SSIM $\uparrow$ & LPIPS $\downarrow$ & \#GS (M) $\downarrow$ & FPS $\uparrow$
& PSNR $\uparrow$ & SSIM $\uparrow$ & LPIPS $\downarrow$ & \#GS (M) $\downarrow$ & FPS $\uparrow$ \\
\midrule
3DGS$^*$ 
& 27.21 & 0.815 & 0.214 & -- & -- 
& 23.14 & 0.841 & 0.183 & -- & -- 
& 29.41 & 0.903 & 0.243 & -- & -- \\

3DGS 
& 27.42 & \cellcolor{bestbg}\textbf{0.813} & \cellcolor{bestbg}\textbf{0.218} & 3.300 & 187.70
& 23.62 & 0.844 & 0.179 & 1.830 & 254.50
& 29.45 & 0.899 & 0.247 & 2.810 & 201.20 \\

RadSplat 
& \cellcolor{bestbg}\textbf{27.45} & 0.811 & 0.223 & 2.184 & 247.80
& 23.61 & \cellcolor{bestbg}\textbf{0.847} & \cellcolor{bestbg}\textbf{0.177} & 1.053 & 396.40
& 29.55 & 0.903 & \cellcolor{bestbg}\textbf{0.243} & 1.515 & 345.30 \\

Compact3DGS 
& 27.01 & 0.797 & 0.247 & 1.429 & 281.10
& 23.35 & 0.830 & 0.202 & 0.833 & 358.90
& 29.73 & 0.901 & 0.258 & 1.044 & 366.20 \\

MaskGaussian  
& 27.43 & 0.812 & 0.224 & 1.228 & \cellcolor{bestbg}\textbf{384.70}
& 23.61 & 0.846 & 0.180 & 0.740 & \cellcolor{bestbg}\textbf{558.30}
& 29.78 & \cellcolor{bestbg}\textbf{0.907} & 0.245 & 0.914 & \cellcolor{bestbg}\textbf{532.40} \\

Ours 
& 27.44 & 0.806 & 0.241 & \cellcolor{bestbg}\textbf{0.951} & 240.25
& \cellcolor{bestbg}\textbf{23.79} & 0.842 & 0.194 & \cellcolor{bestbg}\textbf{0.605} & 335.96
& \cellcolor{bestbg}\textbf{29.81} & 0.905 & 0.247 & \cellcolor{bestbg}\textbf{0.624} & 288.21 \\
\bottomrule
\end{tabular}
}

\vspace{10pt}

\caption{Quantitative comparison of our method with prior work on three datasets. PLY (MB) represents the storage size of point cloud files, and Train denotes training time. \colorbox{bestbg}{Best} results are highlighted in bold with light red background.}
\label{tab:tab2_storage_time}
\setlength{\tabcolsep}{2.4pt}
\renewcommand{\arraystretch}{1.05}
\resizebox{\textwidth}{!}{
\begin{tabular}{lcccccc cccccc cccccc}
\toprule
\multirow{2}{*}{Method}
& \multicolumn{6}{c}{Mip-NeRF 360}
& \multicolumn{6}{c}{Tanks\&Temples}
& \multicolumn{6}{c}{Deep Blending} \\
\cmidrule(lr){2-7} \cmidrule(lr){8-13} \cmidrule(lr){14-19}
& PSNR $\uparrow$ & SSIM $\uparrow$ & LPIPS $\downarrow$ & \#GS (M) $\downarrow$ & PLY (MB) $\downarrow$ & Train $\downarrow$
& PSNR $\uparrow$ & SSIM $\uparrow$ & LPIPS $\downarrow$ & \#GS (M) $\downarrow$ & PLY (MB) $\downarrow$ & Train $\downarrow$
& PSNR $\uparrow$ & SSIM $\uparrow$ & LPIPS $\downarrow$ & \#GS (M) $\downarrow$ & PLY (MB) $\downarrow$ & Train $\downarrow$ \\
\midrule
Compact3DGS
& 27.01 & 0.797 & 0.247 & 1.429 & 327.45 & 26m58s
& 23.35 & 0.830 & 0.202 & 0.833 & 198.10 & 14m44s
& 29.73 & 0.901 & 0.258 & 1.044 & 247.93 & 22m08s \\

MaskGaussian
& 27.43 & \cellcolor{bestbg}\textbf{0.812} & \cellcolor{bestbg}\textbf{0.224} & 1.228 & 287.38 & \cellcolor{bestbg}\textbf{20m21s}
& 23.61 & \cellcolor{bestbg}\textbf{0.846} & \cellcolor{bestbg}\textbf{0.180} & 0.740 & 175.13 & \cellcolor{bestbg}\textbf{11m11s}
& 29.78 & \cellcolor{bestbg}\textbf{0.907} & \cellcolor{bestbg}\textbf{0.245} & 0.914 & 216.24 & \cellcolor{bestbg}\textbf{16m37s} \\

Ours
& \cellcolor{bestbg}\textbf{27.44} & 0.806 & 0.241 & \cellcolor{bestbg}\textbf{0.951} & \cellcolor{bestbg}\textbf{224.96} & 26m14s
& \cellcolor{bestbg}\textbf{23.79} & 0.842 & 0.194 & \cellcolor{bestbg}\textbf{0.605} & \cellcolor{bestbg}\textbf{142.97} & 12m58s
& \cellcolor{bestbg}\textbf{29.81} & 0.905 & 0.247 & \cellcolor{bestbg}\textbf{0.624} & \cellcolor{bestbg}\textbf{147.50} & 20m56s \\
\bottomrule
\end{tabular}
}
\end{table*}

\subsection{Detail-Aware Safeguarding and Post-Pruning Optimization}

Although GNSO preserves the dominant local function subspace, aggressive truncation may discard small Gaussian primitives corresponding to thin structures, high-frequency textures, or occlusion boundaries. To alleviate this issue, we assign each inactive Gaussian a detail score according to its gradient response, visibility, scale, and opacity,
\begin{equation}
d_i
=
g_i^{\gamma}
v_i^{\nu}
s_i^{-\rho}
\alpha_i^{\delta},
\label{eq:detail-score-def}
\end{equation}
where $g_i$, $v_i$, $s_i$, and $\alpha_i$ denote the normalized gradient magnitude, accumulated visibility, effective Gaussian scale, and opacity, respectively. The inverse-scale term favors small Gaussian primitives that are more likely to represent fine geometric structures.

From the GNSO-inactive candidates, we retain the top
$K_{\mathrm{detail}}=\lfloor0.05N\rfloor$ Gaussians ranked by
the detail score.
\begin{equation}
K_{\mathrm{detail}}
=
\left\lfloor
0.05N
\right\rfloor,
\label{eq:detail-recovery-budget}
\end{equation}
yielding the final representative set
\begin{equation}
\mathcal{A}
=
\mathcal{A}_{\mathrm{GNSO}}
\cup
\mathcal{A}_{\mathrm{detail}}.
\label{eq:final-active-set}
\end{equation}
This safeguarding step preserves visually important local structures while maintaining a compact Gaussian representation.

After hard pruning, the retained Gaussian primitives are further optimized with densification disabled, ensuring that the representation remains fixed in size. The post-pruning optimization minimizes
\begin{equation}
\mathcal{L}
=
\mathcal{L}_{\mathrm{3DGS}}
+
\lambda_{\mathrm{MSE}}
\mathcal{L}_{\mathrm{MSE}},
\qquad
\lambda_{\mathrm{MSE}}=2,
\label{eq:postprune-loss-func}
\end{equation}
where $\mathcal{L}_{\mathrm{3DGS}}$ is the original reconstruction objective and $\mathcal{L}_{\mathrm{MSE}}$ is an additional reconstruction loss that compensates for the quality degradation introduced by pruning. Since densification is disabled after pruning, the optimization improves rendering fidelity while keeping the number of Gaussian primitives unchanged, resulting in a compact fixed-cardinality 3DGS representation.

\section{Experiments}
\subsection{Experimental Settings}
\noindent\textbf{Implementation Details.}
All experiments are conducted on a single NVIDIA RTX 4090 GPU for 30,000 training iterations. Standard 3DGS densification is performed until iteration 13,000, followed by RRDM accumulation. Hard pruning is applied at iteration 15,000, after which the retained Gaussian primitives are optimized with densification disabled. Unless otherwise specified, the keep ratio, detail safeguarding ratio, and GNSO energy threshold are set to $0.22$, $0.05$, and $0.95$, respectively. The RRDM dimension is 8, the density mixing coefficient is $0.75$, and the exponential decay factor is $0.995$. A base voxel size of $0.03$ is adopted and is increased to $0.06$ for sparse scenes according to the adaptive partitioning strategy. After pruning, an additional reconstruction loss with weight $\lambda_{\mathrm{MSE}}=2$ is introduced to improve reconstruction quality. Except for the adaptive voxel size, all hyperparameters are shared across all scenes.

\noindent\textbf{Compared Methods.}
We compare QIRF with the original 3DGS and three representative compact 3DGS methods: Compact3DGS, MaskGaussian, and RadSplat. Compact3DGS reduces storage through attribute quantization and entropy coding, MaskGaussian performs adaptive primitive selection using learnable Gaussian masks, and RadSplat improves compactness through rendering-aware Gaussian pruning. Unless otherwise stated, all baseline results are taken from the original papers or reproduced using the official implementations with the recommended settings.

\noindent\textbf{Datasets and Evaluation Metrics.}
Following previous compact 3DGS works, experiments are conducted on three public benchmarks: Mip-NeRF 360, Tanks \& Temples, and Deep Blending. Reconstruction quality is evaluated using PSNR, SSIM, and LPIPS. Representation efficiency is measured by the number of retained Gaussian primitives (\#GS), raw PLY storage size, and rendering speed (FPS). Higher PSNR, SSIM, and FPS indicate better performance, whereas lower LPIPS, Gaussian count, and storage size indicate a more compact representation.

\begin{table*}[!t]
\centering
\setlength{\abovecaptionskip}{4pt}
\setlength{\belowcaptionskip}{4pt}
\setlength{\tabcolsep}{2.6pt}
\renewcommand{\arraystretch}{1.08}
\caption{Quantitative ablation study of our core components on three benchmark datasets. GNSO denotes the generalized natural scene orbital subspace selection module; GNSO+Detail represents our full QIRF pipeline equipped with the detail-aware recovery strategy.}
\label{tab:ablation_gnsodetail}
\resizebox{\textwidth}{!}{
\begin{tabular}{l cccccc cccccc cccccc}
\toprule
\multirow{2}{*}{Method}
& \multicolumn{6}{c}{Mip-NeRF 360}
& \multicolumn{6}{c}{Tanks\&Temples}
& \multicolumn{6}{c}{Deep Blending} \\
\cmidrule(lr){2-7} \cmidrule(lr){8-13} \cmidrule(lr){14-19}
& PSNR $\uparrow$ & SSIM $\uparrow$ & LPIPS $\downarrow$ & \#GS (M) $\downarrow$ & PLY (MB) $\downarrow$ & Train $\downarrow$
& PSNR $\uparrow$ & SSIM $\uparrow$ & LPIPS $\downarrow$ & \#GS (M) $\downarrow$ & PLY (MB) $\downarrow$ & Train $\downarrow$
& PSNR $\uparrow$ & SSIM $\uparrow$ & LPIPS $\downarrow$ & \#GS (M) $\downarrow$ & PLY (MB) $\downarrow$ & Train $\downarrow$ \\
\midrule
GNSO
& 27.18 & 0.7933 & 0.2621 & \cellcolor{bestbg}\textbf{0.932} & \cellcolor{bestbg}\textbf{220.50} & 27m09s
& 23.76 & 0.8363 & 0.2029 & \cellcolor{bestbg}\textbf{0.532} & \cellcolor{bestbg}\textbf{125.80} & 13m53s
& \cellcolor{bestbg}\textbf{29.86} & 0.9042 & 0.2515 & \cellcolor{bestbg}\textbf{0.508} & \cellcolor{bestbg}\textbf{120.20} & 21m47s \\
GNSO+Detail
& \cellcolor{bestbg}\textbf{27.44} & \cellcolor{bestbg}\textbf{0.8060} & \cellcolor{bestbg}\textbf{0.2411} & 0.951 & 224.96 & \cellcolor{bestbg}\textbf{26m14s}
& \cellcolor{bestbg}\textbf{23.79} & \cellcolor{bestbg}\textbf{0.8415} & \cellcolor{bestbg}\textbf{0.1940} & 0.605 & 142.97 & \cellcolor{bestbg}\textbf{12m58s}
& 29.81 & \cellcolor{bestbg}\textbf{0.9049} & \cellcolor{bestbg}\textbf{0.2473} & 0.624 & 147.50 & \cellcolor{bestbg}\textbf{20m56s} \\
\bottomrule
\end{tabular}
}
\end{table*}

\subsection{Experimental Results}

Table~\ref{tab:tab1_basic_metric} reports quantitative comparisons between QIRF and recent compact 3DGS methods on Mip-NeRF 360, Tanks\&Temples, and Deep Blending. Compared with vanilla 3DGS, QIRF substantially reduces the number of Gaussian primitives while maintaining competitive reconstruction quality. Specifically, QIRF achieves the smallest Gaussian count on all three datasets, reducing the primitives by approximately $71.2\%$, $67.0\%$, and $77.8\%$, respectively. Despite this aggressive compression, QIRF achieves the best average PSNR on Tanks\&Temples and Deep Blending, and remains highly competitive on Mip-NeRF 360. These results indicate that function-space compression can effectively remove redundant Gaussian primitives without noticeably degrading pixel-level reconstruction fidelity.

Table~\ref{tab:tab2_storage_time} further compares storage size and training cost with Compact3DGS and MaskGaussian. QIRF achieves the smallest raw PLY size across all datasets, reducing the average storage to $224.96$ MB on Mip-NeRF 360, $142.97$ MB on Tanks\&Temples, and $147.50$ MB on Deep Blending. Compared with MaskGaussian, QIRF reduces PLY storage by $21.7\%$, $18.4\%$, and $31.8\%$ on the three datasets, respectively. Compared with Compact3DGS, QIRF reduces storage by $31.3\%$, $27.8\%$, and $40.5\%$, respectively. Although QIRF is not always the fastest method during training, its training time remains comparable to existing compact 3DGS approaches while providing a significantly more compact representation.

Qualitative comparisons are shown in Figure 2. Compared with vanilla 3DGS, Compact3DGS and MaskGaussian, QIRF preserves scene geometry, object boundaries, and high-frequency textures while using fewer Gaussian primitives. Fine structures, such as bicycle spokes, thin object boundaries, reflective regions, vehicle contours, and indoor decorations, remain visually consistent after pruning. These observations suggest that the proposed GNSO-based function-space selection and detail-aware safeguarding preserve visually important structures under high compression ratios.

In terms of rendering efficiency, QIRF achieves $240.3$, $336.0$, and $288.2$ FPS on Mip-NeRF 360, Tanks\&Temples, and Deep Blending, respectively, as reported in Table~\ref{tab:tab1_basic_metric}. This is faster than vanilla 3DGS, showing that reducing redundant primitives directly improves rasterization efficiency. However, QIRF is slower than highly optimized rendering-oriented methods such as MaskGaussian and Compact3DGS. This suggests that the current implementation primarily focuses on function-space redundancy reduction and compact representation, while further system-level optimization, such as active-Gaussian indexing and sparse rasterization, is needed to fully translate compression into rendering speed.

\begin{figure}[t]
\centering
\includegraphics[width=0.9\columnwidth]{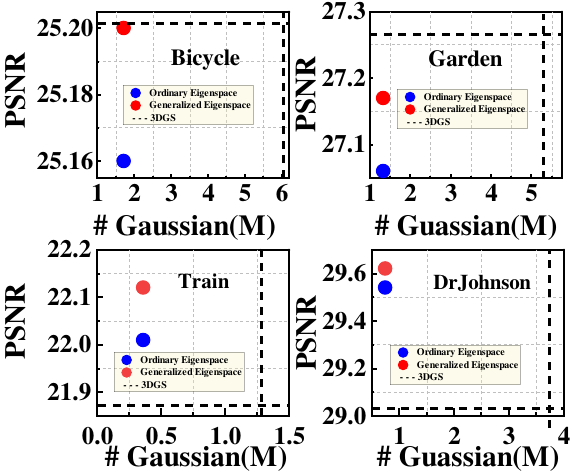}
\caption{PSNR versus the number of retained Gaussians for ordinary and
generalized eigenspace selection. The dashed line represents the
original 3DGS performance.}
\label{fig:eigen_psnr_curve}
\end{figure}
\subsection{Ablation Study}
Compares ordinary eigenspace selection with the proposed generalized eigenspace formulation on Bicycle, Garden, Train, and DrJohnson in Figure 3. Both variants retain a substantially smaller number of Gaussians than the original 3DGS models. However, the generalized formulation consistently achieves higher PSNR under a comparable Gaussian budget. This advantage is particularly evident on Garden and Train, where ordinary eigendecomposition causes a more noticeable quality degradation. The results demonstrate that treating the Gaussian primitives as a non-orthogonal basis and explicitly incorporating the overlap matrix leads to a more informative subspace and more reliable Gaussian selection.

Table~3 shows that detail-aware safeguarding improves reconstruction quality over the GNSO-only model with limited additional storage. On Mip-NeRF 360, PSNR increases from $27.18$ to $27.44$ dB, SSIM from $0.7933$ to $0.8060$, and LPIPS decreases from $0.2621$ to $0.2411$. Similar gains are observed on Tanks\&Temples, while on Deep Blending PSNR remains nearly unchanged but SSIM and LPIPS improve. These gains require only a modest increase in Gaussian count and PLY size, indicating that retaining a small number of structurally important primitives effectively compensates for the quality loss caused by aggressive function-space pruning.

\subsection{Limitations and Future Work}
Several aspects of QIRF remain open for further exploration. The current framework mainly focuses on reducing primitive-level redundancy, while attribute quantization and entropy coding are not incorporated; consequently, the reported storage results primarily reflect raw PLY compression. The adaptive partitioning strategy currently selects from a small set of voxel sizes, and more flexible scene-dependent partitioning may further improve robustness. In addition, the dominant GNSO subspace is mapped back to representative Gaussian primitives rather than directly rendered as a contracted basis. Although QIRF improves rendering efficiency over vanilla 3DGS, further integration with sparse rasterization and dedicated CUDA kernels may provide additional acceleration. These directions offer opportunities to extend QIRF toward more compact and efficient Gaussian radiance-field representations.

\section{Conclusion}
We presented QIRF, a quantum-inspired non-orthogonal function-space compression for 3D gaussian splatting. Rather than treating Gaussian primitives independently, QIRF models neighboring Gaussians as a local non-orthogonal basis and combines an analytic overlap matrix with a Radiance-Response Density Matrix to identify dominant scene subspaces through generalized eigendecomposition. Detail-aware safeguarding further preserves visually important structures under aggressive primitive reduction. Across 13 scenes from three standard datasets, QIRF removes $71.7\%$ of Gaussian primitives on average and achieves approximately $3.54\times$ raw PLY compression while maintaining comparable reconstruction quality, with a marginal average PSNR improvement of $0.10$ dB. These results suggest that non-orthogonal function-space redundancy is an important yet underexplored source of representational redundancy in explicit Gaussian radiance fields. Future work will investigate more flexible spatial partitioning, perceptual response modeling, contracted Gaussian basis optimization, and integration with attribute quantization, entropy coding, and specialized rendering kernels.
\bibliography{References}

\end{document}